%% file: main.tex
\documentclass{article}
\usepackage{iclr2027_conference,times}
\input{math_commands.tex}

\usepackage{mathtools}

\newcommand{\dynsys}{\mathcal{D}}
\newcommand{\init}{\mathbf{I}}
\newcommand{\trans}{\mathbf{F}}
\newcommand{\noise}{\mathbf{E}}
\newcommand{\ctrl}{\mathbf{u}}
\newcommand{\timestep}{\delta}
\newcommand{\horizon}{T}
\newcommand{\reach}{\mathbf{G}}
\newcommand{\avoid}{\mathbf{A}}
\newcommand{\stateset}{\mathbf{X}}
\newcommand{\trajj}{\tau}
\newcommand{\nxt}{\mathit{next}}
\newcommand{\err}{\bm{\epsilon}}

\newcommand{\sens}{h}
\newcommand{\exo}{\bm{\Xi}}
\newcommand{\vxi}{\bm{\xi}}

\usepackage{hyperref}
\usepackage{etoc}
\usepackage{url}
\usepackage{wrapfig}
\usepackage{algorithm}
\usepackage{algpseudocode}
\usepackage{booktabs}
\usepackage{tikz}
\usetikzlibrary{arrows.meta,shapes.geometric,calc}
\usepackage{pgfplots}
\pgfplotsset{compat=1.17}
\usepgfplotslibrary{fillbetween}
\definecolor{clpblue}{RGB}{31,86,148}
\definecolor{clpgray}{RGB}{99,106,114}
\input{figures/aebs_clip}

\title{Stochastic World Models for Verifying \\ Vision-Based Neural Feedback Systems}

\author{I. Samuel Akinwande$^{1}$ \quad Mykel J. Kochenderfer$^{1}$ \quad Clark Barrett$^{2}$ \\
$^{1}$Department of Aeronautics and Astronautics, Stanford University \\
$^{2}$Department of Computer Science, Stanford University \\
\texttt{\{samakin,mykel,barrettc\}@stanford.edu}
}

\iclrfinalcopy
\begin{document}

\maketitle
\etocdepthtag.toc{mtbody}

\begin{abstract}
Verifying a vision-based neural feedback system requires a model of the
observations its controller acts upon. Such a model must capture the variation
the sensor produces, while remaining tractable for closed-loop analysis.
Generative adversarial networks (GANs) have served as perception surrogates,
but they are large, reproduce complex scenes poorly, and are hard to verify.
We explore stochastic world models as a richer class of perception surrogates.
We train a world model with physically grounded latents, built from operations
that standard verifiers bound. It reproduces held-out frames more faithfully
than GAN surrogates with up to 130 times as many parameters. To verify these
surrogates, we develop a procedure that combines falsification, adaptive
refinement, symbolic, and backward analyses. On an emergency braking
benchmark with a GAN surrogate, our procedure resolves the entire state space,
38\% of which the state-of-the-art verifier left unresolved. On the RGB version
of the benchmark, where no verification results have previously been
reported, our procedure resolves over 80\% of the state space with a world
model surrogate.
\end{abstract}
\section{Introduction}
Modern autonomous systems act on uncertain, high-dimensional observations
instead of exact state information. These observations are often images, and
a neural network maps them to control actions, closing the loop between
perception and actuation. Systems in which a neural network closes the
feedback loop are called \emph{neural feedback systems}, and they are
increasingly deployed in safety-critical domains such as aerial
navigation~\citep{kaufmann2023champion}, humanoid
robotics~\citep{radosavovic2024humanoid}, and autonomous
driving~\citep{nebot2026era}. Verification can establish that such a system
is safe before it is deployed, and while mature methods exist for classical
control systems~\citep{mitchell2005time,bansal2017hamilton}, verifying neural
feedback systems remains a challenge. Recent work verifies state-based neural
feedback systems, whose controllers compute actions directly from the
state~\citep{akinwande2025polyhedral,akinwande2026closing,kochdumper2023open,rober2023backward},
but the vision-based case remains largely open. The main obstacle is
modeling the environment. Verifying a vision-based system requires a model
of the sensor and the scene it observes, and the resulting safety guarantee
holds for the real system to the extent that this model is faithful. The
model must be expressive enough to capture the variation in the images the
system will encounter, yet compact enough for a verifier to reason about,
and these requirements often conflict.

Prior work replaces the camera with a generative adversarial network (GAN)
that renders images for each state~\citep{katz2022verification,cai2025scalable},
but existing verifiers leave the resulting surrogate systems partly
unresolved~\citep{cai2025scalable}. Alternative surrogates, including
variational autoencoders and formal perception
models~\citep{parameshwaran2025scalable,hsieh2022verifying}, either fail to
capture the variation in the sensor's images or are intractable to verify in
closed loop. World models learn to generate the observations of an
environment, and failure modes found in them have been shown to transfer to
the real world~\citep{ward2026foundational}. \citet{geng2025deterministic}
use a world model for closed-loop verification, but its architecture is
deterministic. It produces a single observation per state and does
not model the environmental variation that makes vision-based verification
hard.

We claim that careful surrogate design can ease the tension between
expressiveness and tractability. A GAN draws its variation from a noise
vector with no direct physical meaning, so the box of noise values a verifier
bounds corresponds to no clear range of conditions. A deterministic world
model is easier to verify but has no variation to bound. We instead train a
stochastic world model that concentrates the variation in a few latents
grounded in physical quantities (Figure~\ref{fig:surrogates}). We frame a
vision-based neural feedback system as a state-based system whose sensor is
a generative model. To verify the resulting systems, we build a procedure
that combines falsification, adaptive refinement, symbolic, and backward
analyses. We evaluate the world model on aircraft taxiing and emergency
braking case studies, and the procedure on emergency braking. Our
contributions are as follows:
\begin{itemize}
    \item \textbf{Formalization.} A formalization of vision-based neural
    feedback systems as state-based systems whose sensor is a generative
    model over a set of latents. In this formalization, the state-based
    case is the special case where the sensor is the identity map, so
    analysis techniques for state-based systems extend to the vision-based
    setting.
    \item \textbf{Improved Modeling.} A stochastic world model with latents
    grounded in physical quantities, so that the possible observations
    at a state correspond to a box of environmental conditions. The model
    is built from operations that standard verifiers bound, is trained on
    closed-loop rollouts of the controller, and reproduces held-out frames more
    faithfully than the GAN surrogates of both case studies, with up to 130
    times fewer parameters.
    \item \textbf{Verification Procedure.} A verification procedure that
    combines falsification, adaptive refinement, symbolic, and backward
    analyses. On the emergency braking benchmark, it resolves the entire
    state space with the released GAN surrogate, 38\% of which the
    state-of-the-art verifier left unresolved. On the RGB version of the
    benchmark, it resolves over 80\% of the state space with a world model
    surrogate.
\end{itemize}
\section{Related Work}
\label{sec:related}
\noindent\textbf{Verification algorithms.}\quad
A few families of methods bound a network's behavior over an input set. Linear
relaxation-based perturbation analysis (LiRPA)~\citep{xu2020automatic}
propagates linear bounds built from per-node abstractions that are exact at
affine layers and sound at nonlinearities. Mixed-integer encodings represent a
piecewise-linear network exactly, defining a binary variable for each unstable
neuron~\citep{tjeng2019evaluating}. Abstractions may be too loose to decide a
property, so complete verifiers pair abstractions with refinement via branch and
bound~\citep{xu2020fast,wang2021beta}. Refinement partitions the input
set or fixes unstable activations, bounds each subproblem, and verifies the
property on every subproblem, or refutes it once one
subproblem yields a counterexample.
Verifiers for state-based neural feedback systems build on this machinery by
composing the dynamics with the controller. Forward methods over-approximate
the reachable set one step at a time, either by encoding the dynamics and
controller together as a mixed-integer
program~\citep{sidrane2022overt,akinwande2025polyhedral} or by propagating
sets or bounds through their
composition~\citep{kochdumper2023open,akinwande2026closing}. Backward
methods compute the states from which the unsafe set is
reachable~\citep{rober2023backward} or the states that reach the
goal~\citep{akinwande2026fabric}. 

\noindent\textbf{Perception surrogates.}\quad
A perception surrogate stands in for the sensor during verification.
\citet{katz2022verification} replace a camera with a conditional
GAN~\citep{mirza2014conditional} that generates observations from the
low-dimensional state, yielding a map from states to observations that
neural network verifiers can analyze~\citep{julian2019guaranteeing}.
\citet{cai2025scalable} verify this benchmark and introduce an emergency
braking benchmark with grayscale and RGB variants~\citep{zhang2019self}, leaving
38\% of the grayscale system unresolved and the RGB system unverified.
Other surrogates include variational
autoencoders~\citep{parameshwaran2025scalable}, formal models of the
perception pipeline~\citep{santacruz2022nnlander,hsieh2022verifying}, and
deterministic world models~\citep{geng2025deterministic}, which decode a
single observation per state.

\noindent\textbf{World models.}\quad
World models learned from pixels have been applied to
game-playing~\citep{hafner2021mastering} and have since served as
environment surrogates across domains~\citep{hafner2025mastering}, including
driving~\citep{wang2024drivedreamer}. Failures found in world models have
been shown to transfer to real systems~\citep{ward2026foundational}. The
latents of a world model typically carry no physical meaning, so a guarantee
over a set of latents does not say which conditions it covers. Recent work
argues for latents that are physically interpretable by
construction~\citep{peper2025four} and learns such representations under
weak supervision~\citep{mao2026physically}. 

\noindent\textbf{Combined strategies.}\quad
When no single analysis decides a property, prior work combines several.
Refinement tightens bounds at the cost of more subproblems, whether by
partitioning the
input~\citep{everett2021robustness}, splitting only where a candidate
violation survives~\citep{rober2024carv}, or refining wherever a
counterexample proves spurious~\citep{elboher2020abstraction,li2026drgbab}.
Falsification has been paired with reachability so that one analysis
directs the
other~\citep{dreossi2019compositional,dreossi2019verifai,tsujio2025ramponn},
and forward and backward reachability have been integrated into a single
procedure for neural feedback systems~\citep{akinwande2026fabric}. 
\section{Problem Formulation}
\label{sec:formulation}
\noindent\textbf{Notation.}\quad
We write $2^{\stateset}$ for the power set of $\stateset$, $[i..j]$ for
$\{i, \ldots, j\}$, $[n]$ for $[1..n]$, and $S_t$ for the $t$-th element
of a sequence $S$. Functions apply to sets elementwise, with the results
unioned.

\subsection{Neural Feedback Systems}
\label{sec:bg-nfs}
A neural feedback system is a dynamical system controlled by a neural
network. We model the system in discrete time, with a controller that acts
on observations of the state, and represent it by the tuple
$\dynsys = \langle m, n, d, \init, \trans, \noise, \exo, \sens, \ctrl, B,
\timestep, \horizon, \reach, \avoid \rangle$. The state $\vs \in \R^n$
evolves under a vector field $\trans = (f_1, \ldots, f_n)$ with
$f_i : \R^n \to \R$, subject to perturbations drawn from
$\noise \subseteq \R^n$. A sensor $\sens : \R^n \times \exo \to \R^d$ maps
the state and an exogenous input $\vxi \in \exo$ to an
observation $\vo \in \R^d$. The controller
$\ctrl : \R^d \to \R^m$ computes an action from the observation, and the
action drives the dynamics through a controller-input matrix
$B \in \R^{n \times m}$. The system starts in $\init \subseteq \R^n$,
advances in steps of length $\timestep$ for $\horizon$ steps, and is
evaluated against a goal set $\reach \subseteq \R^n$ and time-indexed unsafe
states $\avoid : [0..\horizon] \to 2^{\R^n}$. One step of the closed loop is
\begin{align}
  \nxt^{\dynsys}(\vs)
  &\coloneqq \{ \vs + (\trans(\vs) + B\ctrl(\sens(\vs, \vxi)) + \err)\,
  \timestep \mid \err \in \noise,\ \vxi \in \exo \}.
  \label{eq:next-state}
\end{align}
Starting from $\stateset_0 \subseteq \init$, the trajectory
$\trajj^{\dynsys}(\stateset_0) \coloneqq (\stateset_0, \ldots, \stateset_\horizon)$
with $\stateset_t \coloneqq \nxt^{\dynsys}(\stateset_{t-1})$ for
$t \in [\horizon]$ denotes the states that can be reached at each time step.

\input{figures/surrogate_comparison}
The verification literature has focused on the \emph{state-based} case, where
$d = n$ and $\sens(\vs, \vxi) = \vs$. In that setting, the controller
observes the state exactly, and $\exo$ plays no role. This paper addresses the \emph{vision-based} case, where $\sens$ is a
camera, $d \gg n$, and $\vxi$ denotes environmental factors such as lighting
and weather. Such a sensor has no closed-form description, so we introduce a
\emph{perception surrogate} $g : \R^n \times \sZ \to \R^d$, a generative
model that maps a state and a latent $\vz \in \sZ$ to an
observation. Substituting $g$ for $\sens$ and $\sZ$ for $\exo$ in $\dynsys$
yields the \emph{surrogate system}. This framing places
vision-based systems within the state-based formalism, so state-based
verification techniques extend to them.

\input{figures/reach_avoid}
\noindent\textbf{Reach-Avoid Specifications.}\quad
The system $\dynsys$ is \emph{safe} if every trajectory enters the goal set
at some step and no trajectory intersects the unsafe states at any step:
\begin{align}
  &\forall \vs \in \init.\ \exists t \in [0..\horizon].\
    \trajj^{\dynsys}(\{\vs\})_t \subseteq \reach, \label{eq:reach} \\
  &\forall t \in [0..\horizon].\
    \trajj^{\dynsys}(\init)_t \cap \avoid(t) = \emptyset. \label{eq:avoid}
\end{align}
Equations~\ref{eq:reach} and~\ref{eq:avoid} are the reach and avoid
properties (Figure~\ref{fig:reachavoid}). Since $\nxt^{\dynsys}$ ranges over
every $\vxi \in \exo$, a system satisfying these properties is safe for every environmental condition in $\exo$.

\subsection{Reachability Analysis}
\label{sec:reachability}
Verifying a reach-avoid specification requires sound approximations of the
trajectory $\trajj^{\dynsys}(\stateset_0)$. \emph{Forward} analysis
over-approximates $\trajj^{\dynsys}(\stateset_0)$ and checks it against
$\avoid(t)$ and $\reach$ at each step. \emph{Backward} analysis instead
over-approximates the set of states from which $\avoid$ is reachable within
the horizon and checks that it is disjoint from $\init$, or
under-approximates the set of states that reach $\reach$ and checks that it
contains $\init$. Both analyses accumulate approximation error over the
horizon, since each step starts from the previous step's approximation.
Adaptive refinement~\citep{rober2024carv}, symbolic
analysis~\citep{akinwande2025polyhedral}, and combinations of forward and
backward analysis~\citep{akinwande2026fabric} reduce this error.
\section{Stochastic World Models}
\label{sec:swm}
Our world model is a renderer
$g : \R^n \times \sZ \to \R^{d}$ that
maps a state $\vs$ and a vector of latents $\vz \in \sZ$ to an image.
The state determines the geometry of the scene, and the latents
account for variations between two images at the same state, such as the
lighting. The model has no noise input, so the latents are its only source
of variation. Each latent is a physical quantity ranging over an interval,
so $\sZ$ is a box of physical conditions. The set of images the controller
can observe in state $\vs$,
\begin{equation}
  g(\vs, \sZ) = \{\, g(\vs, \vz) \mid \vz \in \sZ \,\},
  \label{eq:obs-set}
\end{equation}
is then the set of images taken under those conditions. Taking $g$ as the perception surrogate yields a system in which $\vz$ ranges
over $\sZ$ independently at every step.

\noindent\textbf{Architecture.}\quad
The model is a deconvolutional decoder in the style of DCGAN
generators~\citep{radford2016dcgan} and the DreamerV3 image
decoder~\citep{hafner2025mastering}. A linear layer lifts the state to a
coarse feature map, and $L$ transposed-convolution stages double its
resolution in turn:
\begin{align}
  h_0 &= \operatorname{reshape}(W_0 \vs + b_0), \qquad
  h_k = \operatorname{ReLU}\bigl(\operatorname{BN}_k(\operatorname{ConvT}_k(h_{k-1}))\bigr),
  \quad k \in [L].
  \label{eq:decoder}
\end{align}
Latents enter through feature-wise linear modulation
(FiLM)~\citep{perez2018film}, which rescales and shifts a feature map per
channel by an affine function of $\vz$:
\begin{equation}
  \operatorname{FiLM}(h; \vz) = \bigl(1 + \gamma(\vz)\bigr) \odot h + \beta(\vz),
  \qquad (\gamma, \beta)(\vz) = W\vz + b.
  \label{eq:film}
\end{equation}
FiLM is applied once to the last feature map and once to the image before
the output nonlinearity, with a separate projection $(W, b)$ each time:
\begin{equation}
  g(\vs, \vz) = \tanh\Bigl(\operatorname{FiLM}\bigl(
     \operatorname{Conv}\bigl(\operatorname{FiLM}(h_L; \vz)\bigr); \vz\bigr)\Bigr).
  \label{eq:renderer}
\end{equation}
We initialize both projections to zero, so FiLM starts as the identity and
training begins from a state-only decoder, with the latent modulation
learned on top of it. The zero initialization, together with FiLM acting
per channel, keeps the geometry of the image tied to the state and leaves
its appearance to the latents.

\noindent\textbf{Verifiability.}\quad
Apart from the FiLM product, the model uses only linear, convolution, batch
normalization, ReLU, and $\tanh$ layers, all of which standard verifiers
bound. Batch normalization~\citep{ioffe2015batchnorm} is a per-channel
affine map at inference. Layer and group normalization instead compute
their statistics from the input and divide by an input-dependent standard
deviation, a division that verifiers bound loosely. The FiLM product
$\gamma(\vz) \odot h$ multiplies an affine function of $\vz$ by a feature
map that depends on $\vs$, and we bound it with the standard McCormick
relaxation~\citep{mccormick1976computability} for a product of two bounded
quantities.

\noindent\textbf{Training.}\quad
The model is trained by supervised regression on simulated camera images.
We collect closed-loop rollouts in which the controller acts on the true
state, and record each image together with the state $\vs$ and latents
$\vz$ at which it was rendered. The latents are held fixed within a rollout
and varied across rollouts to cover $\sZ$
(Appendix~\ref{app:training}). The objective is a pixel-wise
$\ell_1$ loss plus a structural similarity (SSIM)
term~\citep{wang2004ssim} with equal weights,
$\mathcal{L} = \lVert g(\vs,\vz) - \vo \rVert_1 + 1 -
\mathrm{SSIM}(g(\vs,\vz), \vo)$. The model is fit to the images alone and
is never tuned to the controller or the verification process. It is small
enough to train in under a minute on a single H100.
\input{figures/renderer}
\section{Verification Procedure}
\label{sec:procedure}
We extend state-based verification of neural feedback systems to the
vision-based setting, building on the formalization in
Section~\ref{sec:formulation}. Recent state-based methods combine open- and
closed-loop verification~\citep{kochdumper2023open,akinwande2026closing},
and we take the same approach. We start from an existing open-loop
verifier~\citep{xu2020fast,wang2021beta} and add methods to support
closed-loop analysis. We partition the initial set $\init$ uniformly into
cells and resolve each cell with falsification, forward analysis, adaptive
refinement, symbolic analysis, and backward analysis, which the following
paragraphs describe (Algorithm~\ref{alg:procedure}).

\noindent\textbf{Falsification} is the cheapest part of the procedure, and
in our experiments, uniform sampling of initial states and latents finds
most failing trajectories. For each cell, the falsifier samples initial
states and latents densely, and simulates each sample through the
perception surrogate, the controller, and the dynamics. Trajectories that
enter an unsafe set are reevaluated in higher-precision arithmetic. A cell
with a confirmed unsafe trajectory is \emph{falsified}, and every other
cell is \emph{unresolved}. A falsified cell may still contain safe states,
but we mark the whole cell unsafe so that the verified region remains a
sound under-approximation of the safe set.

\noindent\textbf{Forward analysis} computes per-step bounds on the states
reachable from each unresolved cell. Sound bounds on the next states
require bounds on the observations the perception surrogate can produce
from the current states and any latents in $\sZ$, on the actions the
controller can take on those observations, and on the states the vector
field $\trans$ reaches under those actions. We compute these bounds by
building on existing algorithms and abstractions for verifying neural
networks~\citep{wang2021beta} and state-based neural feedback
systems~\citep{akinwande2026closing}. The result is a sound
over-approximation of the forward trajectories of every state in the
cell. If the over-approximation satisfies the reach-avoid specifications, the cell is \emph{verified};
otherwise, it remains unresolved. To reduce the conservatism of the
over-approximation, we apply \emph{abstraction optimization}, which tunes
the free parameters of each abstraction, such as the slopes of ReLU
relaxations. Our scheme builds on similar schemes for neural
network~\citep{xu2020fast} and state-based
system~\citep{akinwande2026closing} verification.
\begin{algorithm}[t]
\caption{Verification procedure.}
\label{alg:procedure}
\begin{algorithmic}[1]
\Require surrogate system $\dynsys$ with latent box $\sZ$; partition
  resolution $N$; refinement depth $D_A$ for each analysis $A$
\For{each cell $C$ of the uniform partition of $\init$ into $N^n$ cells}
    \If{\Call{Falsify}{$C$}}
        \State mark $C$ \textsc{falsified}
    \Else
        \State $Q \gets \{C\}$
        \For{$A \in (\textsc{Forward}, \textsc{Symbolic}, \textsc{Backward})$}
            \State $Q \gets \Call{Resolve}{A, Q, D_A}$
              \Comment{refined subproblems $A$ leaves unverified}
        \EndFor
        \State mark $C$ \textsc{verified} if $Q = \emptyset$, and
          \textsc{unresolved} otherwise
    \EndIf
\EndFor
\end{algorithmic}
\end{algorithm}

\noindent\textbf{Adaptive refinement} reduces conservatism further by
shrinking the domain over which each abstraction is built, since
abstraction optimization alone can leave large portions of the state
space unresolved. Our scheme combines input
splitting~\citep{rober2024carv}, neuron splitting~\citep{wang2021beta},
and enclosure refinement~\citep{akinwande2026closing}. Each refinement
splits the problem into subproblems, and a cell is verified only if all of
its subproblems are. We refine up to a fixed depth, and subproblems that
remain unverified at that depth are evaluated via symbolic analysis.

\noindent\textbf{Symbolic analysis} keeps the correlations between steps
that per-step forward analysis discards. It unrolls the closed loop over
several steps, and bounds the trajectory in one shot, minimizing the
approximation error that accumulates over the horizon
(Section~\ref{sec:reachability}), and that abstraction optimization and
refinement often cannot recover. Symbolic analysis was developed for
state-based systems~\citep{sidrane2022overt}, and has since been extended
to the vision-based setting~\citep{cai2025scalable}. We build on the
vision-based method with optimizations that make the analysis tractable
for our surrogates. Symbolic analysis remains expensive, so we reserve it
for the cells that the previous analyses leave unresolved.

\noindent\textbf{Backward analysis} over-approximates the set of states
from which the closed loop reaches the unsafe set $\avoid$. Forward
analysis must propagate its over-approximation over the full horizon, and
its abstractions loosen as that set grows, so some of its conservatism is
inherent to its direction. For some problems, the backward computation is
tighter. Recent work combines backward and forward analysis to verify
reach-avoid specifications~\citep{akinwande2026fabric}, and we extend this
idea to the vision-based setting. We carry the forward bound to an intermediate step,
and use backward analysis from $\avoid$ to show that the forward set at
that step cannot reach $\avoid$ within the remaining horizon. This
combination resolves cells that forward analysis alone cannot.
\section{Evaluations}
We evaluate on two vision-based control benchmarks from the surrogate
verification literature. On both, we measure the fidelity of the world
model against held-out images (Section~\ref{sec:res-modeling}). On
emergency braking, the benchmark that remains open, we also run our
verification procedure (Section~\ref{sec:res-verification}). All bound
computations run on an NVIDIA H100, and we report cost in GPU-hours.
\subsection{Case studies}
\label{sec:case-studies}
\paragraph{Aircraft Taxiing.}
We wish to verify that an aircraft taxiing down a runway stays on the
runway. The state is the aircraft's crosstrack position $p$ and heading
error $\theta$, which evolve under a nonlinear map. A camera mounted on
the wing captures images, a perception network reads them to estimate the state, and a proportional controller uses the
estimate to set the steering angle.

\noindent\textbf{Baselines.}\quad
Authors in \citet{katz2022verification} replace the camera with a conditional GAN
(cGAN) trained on X-Plane images, with a two-dimensional latent that
covers variation at a fixed state. To keep the resulting system verifiable,
they distill the GAN into a smaller MLP. They partition the initial set
into $128 \times 128$ cells and verify the surrogate system with standard
neural network verification tools. \citet{cai2025scalable} have since
verified the benchmark as well, so we focus on the modeling problem.
Appendix~\ref{app:nfs} gives further details on the benchmark.

\paragraph{Automatic Emergency Braking (AEBS).}
We wish to verify that an autonomous vehicle stops short of a stationary
obstacle. The state is the distance to the obstacle and the vehicle's
speed. A front-facing camera captures images, a perception network reads
them to estimate the distance, and a controller trained with deep
deterministic policy gradient (DDPG) maps the estimate and the true speed
to a braking command at 20\,Hz.

\noindent\textbf{Baselines.}\quad
Authors in \citet{cai2025scalable} replace the camera with a GAN and check the
property for every state in each cell of a $100 \times 100$ grid over the
initial set and for every image the surrogate can render. They test a
grayscale cGAN with a convolutional perception head and an RGB
self-attention GAN (SAGAN) with an attention head. Their approach leaves
38\% of the state space unresolved with the cGAN and reports no results
for the SAGAN at 20\,Hz.
\subsection{Modeling Quality}
\label{sec:res-modeling}
\input{figures/fidelity_strips}
\textbf{Our world models reproduce held-out frames more faithfully than
the GAN surrogates on both case studies.} Since a verification result
covers the set of images the surrogate can render, a higher-fidelity
surrogate makes that result more informative about the system it stands
in for. We measure fidelity on held-out frames by root mean square error
(RMSE) and by structural similarity (SSIM), and report the results in
Table~\ref{tab:fidelity}. On Aircraft Taxiing, even our smallest world
model, with 50k parameters, renders the scene better than both the
2.76M-parameter DCGAN and the 232k-parameter MLP GAN distilled from it,
and fidelity improves with model size. On AEBS, our world models reach
higher fidelity than GANs with 14 to 130 times as many parameters. The gap
is largest in SSIM, which measures whether the structure of the scene is
preserved (0.713 and 0.823 for our models against 0.449 and 0.489 for
the GANs). The GANs reproduce the mean intensity of the scene but lose its
structure (Figure~\ref{fig:fidelity}). The driving scene is more complex
than the runway, and the modeling errors are correspondingly larger.

\begin{table}[t]
  \centering
  \small
  \caption{Surrogate fidelity on held-out frames. The MLP GAN is distilled
  from the DCGAN. Bold marks the best value for each benchmark.}
  \label{tab:fidelity}
  \begin{tabular}{llrrr}
    \toprule
    Benchmark & Surrogate & Params & RMSE $\downarrow$ & SSIM $\uparrow$ \\
    \midrule
    Aircraft Taxiing & DCGAN~\citep{katz2022verification}& 2.76M & 0.0414 & 0.855 \\
                     & MLP GAN~\citep{katz2022verification}& 232k  & 0.0407 & 0.849 \\
                     & World model (ours)    & 50k   & 0.0368 & 0.867 \\
                     & World model (ours)    & 136k  & 0.0330 & 0.898 \\
                     & World model (ours)    & 231k  & \textbf{0.0320} & \textbf{0.904} \\
    \midrule
    AEBS (gray)      & cGAN~\citep{cai2025scalable}& 430k  & 0.1862 & 0.449 \\
                     & World model (ours)    & 30k   & \textbf{0.1667} & \textbf{0.713} \\
    \midrule
    AEBS (RGB)       & SAGAN~\citep{cai2025scalable}& 3.86M & 0.1908 & 0.489 \\
                     & World model (ours)    & 29k   & \textbf{0.1181} & \textbf{0.823} \\
    \bottomrule
  \end{tabular}
\end{table}
\subsection{Verification results}
\label{sec:res-verification}
\paragraph{Our verification procedure completely solves the
grayscale AEBS benchmark.} Table~\ref{tab:cgan} compares our procedure
with the method of \citet{cai2025scalable}. Their approach reaches no
conclusion on 38\% of the state space, while ours resolves every cell. We
falsify 3{,}464 cells, one more than their method finds. Of the 6{,}536
cells we verify, forward analysis verifies 2{,}208, abstraction
optimization and adaptive refinement 551, symbolic analysis 3, and
backward analysis 3{,}774, including all 2{,}669 cells they verify
(Table~\ref{tab:cgan-methods}, Figure~\ref{fig:cellmap}). Verifying the benchmark cost 1{,}043
GPU-hours, about 1{,}030 of them on the cells their method could not
resolve. The 2{,}669 cells they verify take us under 10 GPU-hours. They
report 374 GPU-hours for the whole benchmark, on hardware they do not
specify.
\begin{table}[b]
\centering
\caption{Verification results on the 10{,}000 cells of the AEBS
benchmark. Bold marks the better verified and unresolved counts on the
cGAN. On the world model, 1{,}822 unresolved cells are not yet analyzed.}
\label{tab:cgan}
\small
\begin{tabular}{ll|r|rrr}
\toprule
Surrogate & Verifier & GPU-hours & Verified & Falsified & Unresolved \\
\midrule
cGAN (grayscale) & \citet{cai2025scalable} & 374 & 2{,}669 & 3{,}463 & 3{,}868 \\
cGAN (grayscale) & Ours & 1{,}043 & \textbf{6{,}536} & 3{,}464 & \textbf{0} \\
\midrule
World model (RGB) & Ours & 550 & 4{,}677 & 3{,}498 & 1{,}825 \\
\bottomrule
\end{tabular}
\end{table}
\paragraph{Our verification procedure resolves over 80\% of the RGB AEBS
benchmark.} \citet{cai2025scalable} report no results for their SAGAN RGB
surrogate. In our experiments, the verification problem the SAGAN induces
does not fit in the memory of an H100, and we are not aware of any
verification procedure that can handle it. Our RGB world model reproduces
held-out frames more faithfully than the SAGAN (Table~\ref{tab:fidelity}),
and with it as the surrogate, our procedure resolves over 80\% of the state
space in 550 GPU-hours. Of the 1{,}825 unresolved cells, 1{,}822 have not
yet been analyzed. On the grayscale benchmark, our procedure resolved over
80\% of the cells in 160 GPU-hours and spent the remaining 880 on the last
20\%. If the RGB benchmark follows the same pattern, verifying it fully
will take about 3{,}600 GPU-hours, roughly the time \citet{cai2025scalable}
report for verifying most of a simpler 10\,Hz version of the RGB
benchmark.
\input{figures/cellmap}
\paragraph{Verifying attention-based perception heads remains a challenge.}
The RGB result in Table~\ref{tab:cgan} uses a convolutional perception head
in place of the attention head of the released benchmark, since efficient
abstractions for attention remain an open problem. Our procedure can
verify systems with attention-based heads, but at a higher cost.
Table~\ref{tab:budget} reports how many cells our procedure verifies
within 10 GPU-hours on the RGB world model for three perception heads
(Appendix~\ref{app:settings}). The convolutional head yields more
than twice as many verified cells as an attention head of similar size (120
against 55). The attention head at the released size, with 1.73M
parameters, is also verifiable, but only 10 cells are verified in the same
budget. However, the larger head reads distances from held-out frames more than twice
as accurately.

\begin{table}[t]
\centering
\caption{Cells verified within 10 GPU-hours on the RGB world model,
starting from a $10 \times 10$ block, using various perception heads. Error is the
mean absolute distance error on held-out frames.}
\label{tab:budget}
\small
\begin{tabular}{lrrrr}
\toprule
Perception head & Params & Verified & GPU-hours & Error (m) \\
\midrule
Convolutional              & 98k   & 120 & 9.75 & 0.384 \\
Attention, small           & 112k  & 55  & 9.91 & 0.383 \\
Attention, released size   & 1.73M & 10  & 9.67 & 0.166 \\
\bottomrule
\end{tabular}
\end{table}
\section{Conclusions and Future Work}
We investigated methods for verifying vision-based neural feedback systems through
perception surrogates. We formalized such a system as a state-based system
whose sensor is a generative model, and extended state-based verification techniques to it. We then trained stochastic world models
with grounded latents, and on both case studies they reproduce held-out frames more faithfully than the GAN surrogates of
prior work, despite having up to 130 times fewer parameters. To verify the resulting
systems, we built a procedure that combines falsification, forward analysis
with abstraction optimization and adaptive refinement, symbolic analysis,
and backward analysis. On the grayscale emergency braking benchmark, the
procedure resolves every cell, including the 38\% that the
state-of-the-art verifier left unresolved. On the previously unsolved RGB benchmark, it resolves
over 80\% of the state space using a world model surrogate.

\textbf{Limitations.} Our guarantees hold for the surrogate
system, and our ability to transfer the guarantees depends on the fidelity of our surrogates. 
Our world models are trained and
evaluated on simulated frames. We measure their fidelity empirically, and no
formal guarantee relates the surrogate to the camera it replaces. Our RGB
results are for a convolutional perception head, and we have not verified
the benchmark with the released attention head.
Finally, symbolic analysis contributes little. It decides three cells on the grayscale benchmark and none so far on the RGB benchmark, and we have not yet
determined why.

\textbf{Next steps.} Tighter abstractions for attention would let us verify
the released perception head, and world models trained on real camera data
would test how far our fidelity results transfer. We also plan to find out
why symbolic analysis helps so little, and to study harder, more nonlinear
benchmarks and richer specifications.

\newpage
\bibliography{refs}
\bibliographystyle{iclr2027_conference}

\clearpage
\appendix
\etocdepthtag.toc{mtappendix}
\etocsettagdepth{mtbody}{none}
\etocsettagdepth{mtappendix}{subsection}
\etocsettocstyle{\section*{Appendix Contents}}{}
\tableofcontents
\input{appendix}

\end{document}

%% file: math_commands.tex
\usepackage{amsmath,amsfonts,bm}

\def\eqref#1{equation~\ref{#1}}

\def\1{\bm{1}}

\def\vzero{{\bm{0}}}

\def\va{{\bm{a}}}

\def\vo{{\bm{o}}}

\def\vs{{\bm{s}}}

\def\vz{{\bm{z}}}

\def\mI{{\bm{I}}}

\DeclareMathAlphabet{\mathsfit}{\encodingdefault}{\sfdefault}{m}{sl}
\SetMathAlphabet{\mathsfit}{bold}{\encodingdefault}{\sfdefault}{bx}{n}

\def\gN{{\mathcal{N}}}

\def\sZ{{\mathbb{Z}}}

\newcommand{\R}{\mathbb{R}}

%% file: figures/aebs_clip.tex
\definecolor{skyday}{RGB}{74,134,196}
\definecolor{skydusk}{RGB}{70,78,96}
\definecolor{skyhaze}{RGB}{186,194,204}
\definecolor{horday}{RGB}{226,236,244}
\definecolor{hordusk}{RGB}{158,160,168}
\definecolor{horhaze}{RGB}{222,226,230}
\definecolor{asphalt}{RGB}{104,106,110}
\definecolor{asphdusk}{RGB}{62,64,70}
\definecolor{carbody}{RGB}{47,60,84}
\definecolor{brakered}{RGB}{178,46,38}

%% file: figures/surrogate_comparison.tex
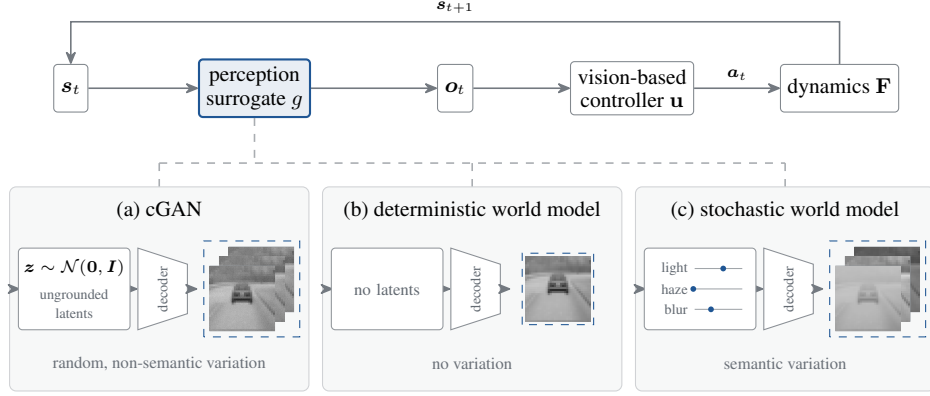
\begin{figure}[t]
\centering
\begin{tikzpicture}[
  scale=0.9, transform shape,
  font=\small,
  >={Stealth[round,length=2.0mm,width=1.6mm]},
  box/.style={draw=clpgray!80, rounded corners=1.5pt, fill=white, inner sep=3pt,
              minimum height=7mm, align=center},
  surr/.style={draw=clpblue, fill=clpblue!10, rounded corners=1.5pt, thick,
               inner sep=3pt, minimum height=7mm, align=center},
  flow/.style={->, draw=clpgray!90, semithick},
  hint/.style={draw=clpgray!55, dashed, semithick},
  panel/.style={draw=clpgray!30, rounded corners=2.5pt, fill=clpgray!5},
  glyph/.style={draw=clpgray!70, rounded corners=1.5pt, fill=white,
                minimum width=16mm, minimum height=12mm, inner sep=2pt,
                text width=15mm, align=center, font=\scriptsize},
  dec/.style={trapezium, trapezium angle=72, shape border rotate=270,
              draw=clpgray!75, fill=white, minimum height=6mm,
              minimum width=12mm, inner sep=1pt, font=\scriptsize},
  bnd/.style={draw=clpblue, dashed, thin},
  tile/.style={inner sep=0pt, draw=clpgray!70, line width=0.3pt},
  tag/.style={font=\scriptsize, align=center, text=clpgray!95},
]

\node[box]  (st) at (1.05,0)  {$\vs_t$};
\node[surr] (g)  at (3.75,0)  {perception\\surrogate $g$};
\node[box]  (ot) at (6.70,0)  {$\vo_t$};
\node[box]  (pi) at (9.30,0)  {vision-based\\controller $\ctrl$};
\node[box]  (f)  at (12.35,0) {dynamics $\trans$};

\draw[flow] (st) -- (g);
\draw[flow] (g)  -- (ot);
\draw[flow] (ot) -- (pi);
\draw[flow] (pi) -- (f) node[midway, above, font=\scriptsize] {$\va_t$};
\draw[flow] (f.north) -- ++(0,0.62) -| (st.north)
      node[pos=0.25, above, font=\scriptsize] {$\vs_{t+1}$};

\draw[hint] (g.south) -- (3.75,-1.10);
\draw[hint] (2.35,-1.10) -- (11.55,-1.10);
\draw[hint] (2.35,-1.10) -- (2.35,-1.45);
\draw[hint] (6.95,-1.10) -- (6.95,-1.45);
\draw[hint] (11.55,-1.10) -- (11.55,-1.45);

\foreach \i/\cx/\ttl in {1/2.35/{(a)~cGAN}, 2/6.95/{(b)~deterministic world model},
                         3/11.55/{(c)~stochastic world model}} {
  \node[panel, minimum width=44mm, minimum height=30mm] at (\cx,-2.95) {};
  \node[font=\small] at (\cx,-1.80) {\ttl};
  \node[dec] (d\i) at (\cx+0.05,-2.95) {};
  \node[font=\tiny, rotate=90, text=clpgray] at (\cx+0.05,-2.95) {decoder};
  \draw[flow] (\cx-2.16,-2.95) -- (\cx-2.05,-2.95);
  \draw[flow] (\cx-0.45,-2.95) -- (d\i.west);
  \draw[flow] (d\i.east) -- (\cx+0.56,-2.95);
}

\node[glyph] (za) at (1.10,-2.95) {$\vz\sim\gN(\vzero,\mI)$\\[3pt]
      {\tiny\textcolor{clpgray}{ungrounded latents}}};
\node[tile] at (3.85,-2.81) {\includegraphics[width=0.95cm, height=0.95cm]
      {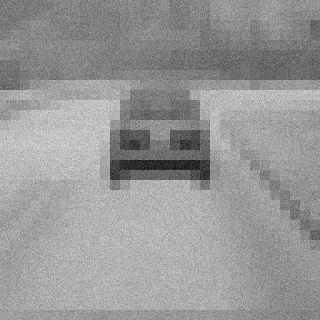}};
\node[tile] at (3.71,-2.95) {\includegraphics[width=0.95cm, height=0.95cm]
      {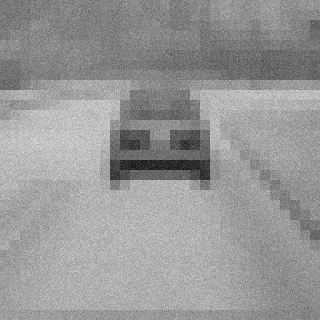}};
\node[tile] at (3.57,-3.09) {\includegraphics[width=0.95cm, height=0.95cm]
      {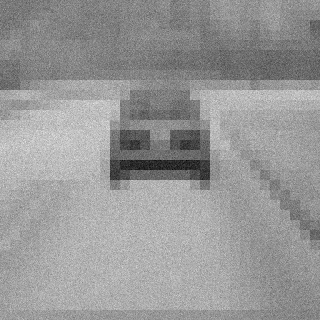}};
\draw[bnd] (3.005,-3.655) rectangle (4.415,-2.245);
\node[tag] at (2.35,-4.05) {random, non-semantic variation};

\node[glyph] (zb) at (5.70,-2.95) {\textcolor{clpgray}{no latents}};
\node[tile] at (8.21,-2.95) {\includegraphics[width=0.95cm, height=0.95cm]
      {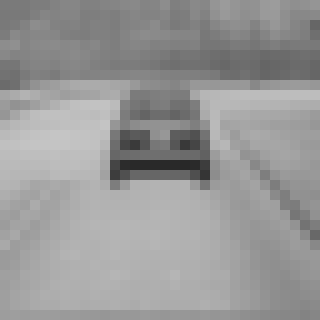}};
\draw[bnd] (7.685,-3.475) rectangle (8.735,-2.425);
\node[tag] at (6.95,-4.05) {no variation};

\node[glyph] (zc) at (10.30,-2.95) {};
\foreach \lbl/\yy/\px in {light/0.30/0.34, haze/0.00/-0.10, blur/-0.30/0.16} {
  \node[font=\tiny, text=clpgray, anchor=east, inner sep=1pt]
        at ($(zc.center)+(-0.14,\yy)$) {\lbl};
  \draw[clpgray!60] ($(zc.center)+(-0.08,\yy)$) -- ($(zc.center)+(0.62,\yy)$);
  \node[circle, fill=clpblue, inner sep=0pt, minimum size=2.4pt]
        at ($(zc.center)+(\px,\yy)$) {};
}
\node[tile] at (13.05,-2.81) {\includegraphics[width=0.95cm, height=0.95cm]
      {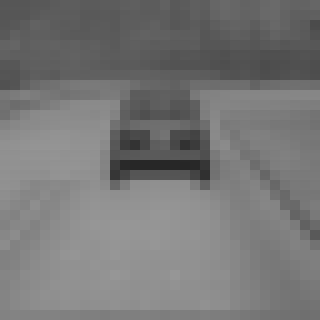}};
\node[tile] at (12.91,-2.95) {\includegraphics[width=0.95cm, height=0.95cm]
      {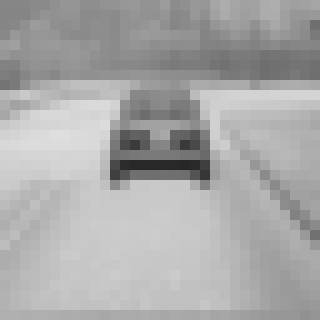}};
\node[tile] at (12.77,-3.09) {\includegraphics[width=0.95cm, height=0.95cm]
      {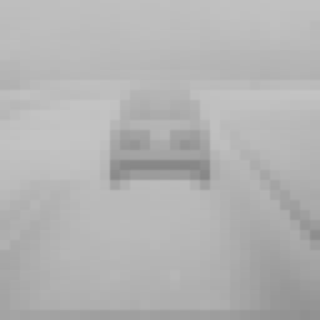}};
\draw[bnd] (12.205,-3.655) rectangle (13.615,-2.245);
\node[tag] at (11.55,-4.05) {semantic variation};

\end{tikzpicture}
\caption{Perception surrogates for vision-based neural feedback systems. The
surrogate $g$ supplies the observation $\vo_t$ the controller acts upon, and so
determines the variation the closed loop can see. (a)~A cGAN draws its
variation from a Gaussian latent with no direct physical meaning. (b)~A
deterministic world model has no latent and no variation. (c)~Our stochastic
world model draws its variation from a few physically grounded latents, whose
ranges form a box that verifiers can bound.}
\label{fig:surrogates}
\end{figure}

%% file: figures/reach_avoid.tex
\definecolor{ragoal}{RGB}{38,118,86}
\definecolor{raunsafe}{RGB}{176,52,44}
\definecolor{raback}{RGB}{193,141,24}
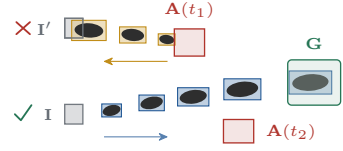
\begin{wrapfigure}{r}{0.36\textwidth}
\vspace{-1.0em}
\centering
\begin{tikzpicture}[
  font=\tiny,
  >={Stealth[round,length=1.4mm,width=1.2mm]},
  tube/.style={draw=clpblue, fill=clpblue!28, line width=0.4pt},
  pre/.style={draw=raback, fill=raback!45, fill opacity=0.6,
             line width=0.4pt},
  truth/.style={fill=black!82},
  initial/.style={draw=clpgray!90, fill=clpgray!22, line width=0.5pt},
  goal/.style={draw=ragoal, fill=ragoal!30, fill opacity=0.3,
               line width=0.5pt, rounded corners=1.5pt},
  unsafe/.style={draw=raunsafe, fill=raunsafe!30, fill opacity=0.42,
                 line width=0.5pt},
]
\draw[initial] (0.10,1.44) rectangle (0.34,1.70);
\node[clpgray, anchor=east] at (0.06,1.57) {$\init'$};
\draw[raunsafe, line width=0.7pt, line cap=round]
      (-0.54,1.49) -- (-0.36,1.65);
\draw[raunsafe, line width=0.7pt, line cap=round]
      (-0.54,1.65) -- (-0.36,1.49);
\draw[initial] (0.10,0.30) rectangle (0.34,0.56);
\node[clpgray, anchor=east] at (0.06,0.43) {$\init$};
\draw[ragoal, line width=0.7pt, line cap=round, line join=round]
      (-0.56,0.43) -- (-0.48,0.35) -- (-0.34,0.55);
\draw[tube] (0.590,0.394) rectangle (0.850,0.566);
\draw[tube] (1.082,0.476) rectangle (1.418,0.684);
\draw[tube] (1.593,0.539) rectangle (2.007,0.781);
\draw[tube] (2.205,0.621) rectangle (2.695,0.899);
\draw[tube] (3.068,0.697) rectangle (3.632,1.003);
\fill[truth] (0.72,0.48) ellipse [x radius=0.120, y radius=0.070, rotate=14];
\fill[truth] (1.25,0.58) ellipse [x radius=0.150, y radius=0.080, rotate=12];
\fill[truth] (1.80,0.66) ellipse [x radius=0.180, y radius=0.090, rotate=8];
\fill[truth] (2.45,0.76) ellipse [x radius=0.210, y radius=0.100, rotate=7];
\fill[truth] (3.35,0.85) ellipse [x radius=0.240, y radius=0.110, rotate=4];
\draw[pre] (1.337,1.345) rectangle (1.563,1.495);
\draw[pre] (0.825,1.373) rectangle (1.175,1.587);
\draw[pre] (0.192,1.406) rectangle (0.648,1.674);
\fill[truth] (1.45,1.42) ellipse [x radius=0.100, y radius=0.060, rotate=-8];
\fill[truth] (1.00,1.48) ellipse [x radius=0.150, y radius=0.080, rotate=-6];
\fill[truth] (0.42,1.54) ellipse [x radius=0.190, y radius=0.095, rotate=-4];
\draw[clpgray!90, line width=0.5pt] (0.10,1.44) rectangle (0.34,1.70);
\draw[->, raback, line width=0.4pt] (1.46,1.13) -- (0.62,1.13);
\draw[->, clpblue!75, line width=0.4pt] (0.62,0.13) -- (1.46,0.13);
\draw[unsafe] (1.55,1.20) rectangle (1.95,1.56);
\node[raunsafe, anchor=south] at (1.75,1.58) {$\avoid(t_1)$};
\draw[unsafe] (2.20,0.05) rectangle (2.60,0.36);
\node[raunsafe, anchor=west] at (2.64,0.20) {$\avoid(t_2)$};
\draw[goal] (3.05,0.55) rectangle (3.75,1.15);
\node[ragoal, anchor=south] at (3.40,1.17) {$\reach$};
\end{tikzpicture}
\caption{Reach-avoid verification. Forward analysis encloses the states
reachable from $\init$ (black) in boxes (blue) that miss each $\avoid(t)$ and
end in $\reach$, so $\init$ is safe. Backward analysis encloses the states
that reach $\avoid(t_1)$ (gold), and these states meet $\init'$, so $\init'$
is unsafe.}
\label{fig:reachavoid}
\vspace{-1.9em}
\end{wrapfigure}

%% file: figures/renderer.tex
\begin{figure}[t]
\centering
\newcommand{\fmap}[4]{%
  \foreach \o in {0.12,0.06,0}
    {\draw[fmapsty] (#1-#3+\o,#2-#3+\o) rectangle (#1+#3+\o,#2+#3+\o);}
  \node[font=\scriptsize, anchor=north, text=clpgray]
        at (#1+0.06,#2-#3-0.10) {#4};}
\begin{tikzpicture}[
  font=\small,
  >={Stealth[round,length=2.0mm,width=1.6mm]},
  box/.style={draw=clpgray!80, rounded corners=1.5pt, fill=white, inner sep=3pt,
              minimum height=7mm, align=center, font=\scriptsize},
  film/.style={draw=clpblue, fill=clpblue!12, rounded corners=1.5pt, thick,
               inner sep=3pt, minimum height=7mm, align=center, font=\scriptsize},
  lat/.style={draw=clpblue!75, rounded corners=1.5pt, fill=white,
              minimum width=15mm, minimum height=10mm, inner sep=2pt},
  fmapsty/.style={draw=clpgray!75, fill=clpgray!10, line width=0.4pt},
  flow/.style={->, draw=clpgray!90, semithick},
  latflow/.style={->, draw=clpblue, semithick},
  tag/.style={font=\scriptsize, text=clpgray!95, align=center},
]

\node[box] (s) at (0.55,0) {$\vs$};
\node[box] (lift) at (1.85,0) {linear\\reshape};
\fmap{3.10}{0}{0.22}{$h_0$}
\fmap{4.20}{0}{0.32}{$h_1$}
\fmap{5.50}{0}{0.44}{$h_k$}
\fmap{6.95}{0}{0.58}{$h_L$}
\node[film] (f1) at (8.25,0) {FiLM};
\node[box]  (cv) at (9.25,0) {Conv};
\node[film] (f2) at (10.25,0) {FiLM};
\node[box]  (th) at (11.20,0) {$\tanh$};
\node[inner sep=0pt, draw=clpgray!70, line width=0.3pt] at (12.70,0)
      {\includegraphics[width=1.05cm, height=1.05cm]
        {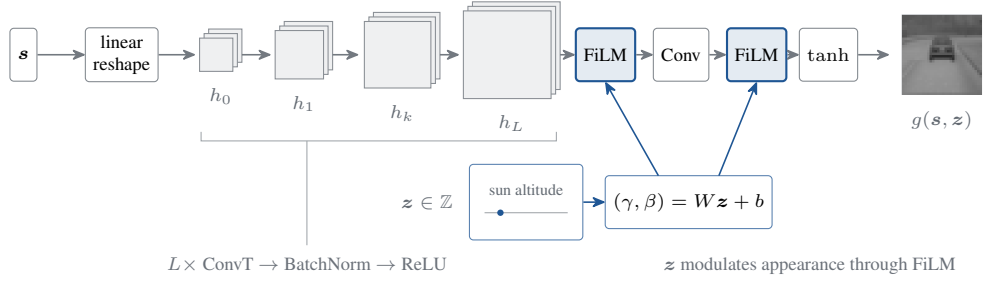}};
\node[font=\scriptsize, text=clpgray, anchor=north] at (12.70,-0.62)
      {$g(\vs,\vz)$};

\draw[flow] (s) -- (lift);
\draw[flow] (lift) -- (2.82,0);
\draw[flow] (3.44,0) -- (3.82,0);
\draw[flow] (4.64,0) -- (4.98,0);
\draw[flow] (6.06,0) -- (6.31,0);
\draw[flow] (7.65,0) -- (f1.west);
\draw[flow] (f1) -- (cv);
\draw[flow] (cv) -- (f2);
\draw[flow] (f2) -- (th);
\draw[flow] (th) -- (12.00,0);

\draw[clpgray!60] (2.90,-1.02) -- (2.90,-1.14) -- (7.60,-1.14) -- (7.60,-1.02);
\draw[clpgray!60] (4.30,-1.14) -- (4.30,-2.52);
\node[tag, anchor=north] at (4.30,-2.55)
      {$L\times$ ConvT $\to$ BatchNorm $\to$ ReLU};

\node[lat] (z) at (7.20,-1.95) {};
\node[font=\tiny, text=clpgray] at ($(z.center)+(0,0.16)$) {sun altitude};
\draw[clpgray!60] ($(z.center)+(-0.55,-0.14)$) -- ($(z.center)+(0.55,-0.14)$);
\node[circle, fill=clpblue, inner sep=0pt, minimum size=2.4pt]
      at ($(z.center)+(-0.34,-0.14)$) {};
\node[font=\scriptsize, text=clpgray, anchor=east] at (6.35,-1.95) {$\vz\in\sZ$};
\node[box, draw=clpblue!75] (aff) at (9.35,-1.95)
      {$(\gamma,\beta) = W\vz + b$};
\draw[latflow] (z) -- (aff);
\draw[latflow] (8.95,-1.60) -- (f1.south);
\draw[latflow] (9.75,-1.60) -- (f2.south);
\node[tag, anchor=north] at (10.95,-2.55)
      {$\vz$ modulates appearance through FiLM};

\end{tikzpicture}
\caption{The world model renderer $g$. A linear layer lifts the state $\vs$
to a coarse feature map, and $L$ stages of transposed convolution, batch
normalization, and ReLU upsample it. The latents $\vz$ act through two FiLM
modulations, which rescale and shift the last feature map and the image.}
\label{fig:renderer}
\end{figure}

%% file: figures/fidelity_strips.tex
\definecolor{fidblue}{HTML}{0072B2}
\definecolor{fidorange}{HTML}{D55E00}
\begin{figure}[t]
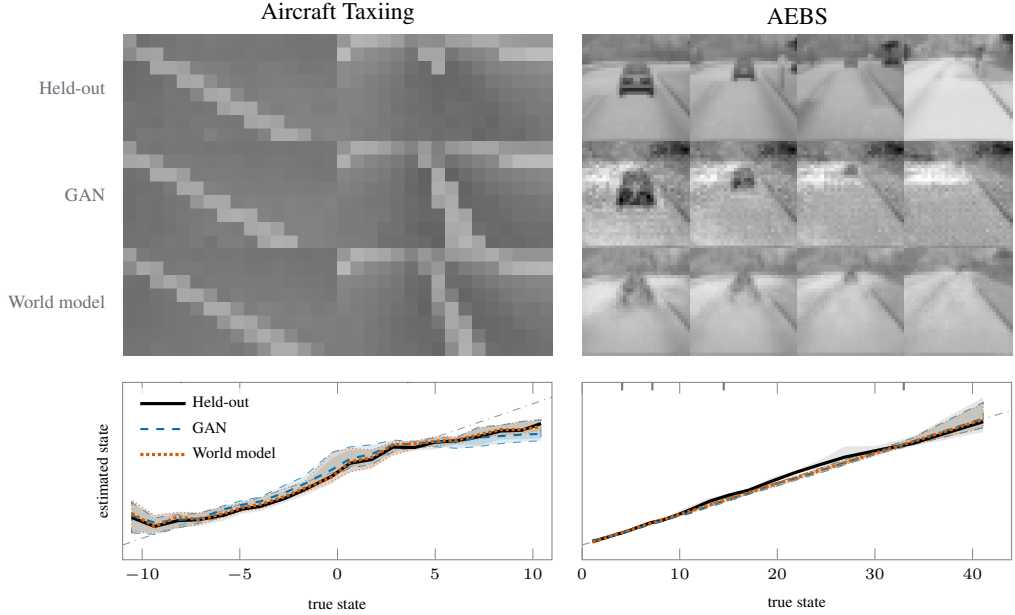

\centering
\begin{tikzpicture}[x=1cm,y=1cm, lab/.style={font=\scriptsize, text=clpgray!95}]
  \def\h{1.42} \def\hh{2.84} \def\g{0}
  \foreach \r/\src in {0/0, 1/1, 2/3}{
    \foreach \c/\col in {0/0, 1/2}{
      \node[anchor=north west, inner sep=0pt] (t\r\c)
        at ({\c*(\hh+\g)+\c*\g}, {-\r*(\h+\g)})
        {\includegraphics[width=\hh cm, height=\h cm]{figures/fidelity/taxinet_r\src_c\col.png}};
    }
  }
  \pgfmathsetmacro{\ax}{2*\hh+3*\g+0.40}
  \foreach \r/\src in {0/1, 1/2, 2/3}{
    \foreach \c in {0,...,3}{
      \node[anchor=north west, inner sep=0pt] (a\r\c)
        at ({\ax+\c*(\h+\g)}, {-\r*(\h+\g)})
        {\includegraphics[width=\h cm, height=\h cm]{figures/fidelity/aebs_gray_r\src_c\c.png}};
    }
  }
  \foreach \r/\lab in {0/{Held-out}, 1/{GAN}, 2/{World model}}
    \node[lab, anchor=east] at ([xshift=-3pt]t\r0.west) {\lab};
  \node[font=\small, anchor=south] at ($(t00.north west)!0.5!(t01.north east)$)
    {Aircraft Taxiing};
  \node[font=\small, anchor=south] at ($(a00.north west)!0.5!(a03.north east)$)
    {AEBS};
  \pgfplotsset{
    fidaxis/.style={scale only axis, anchor=north west, height=2.35cm,
      font=\scriptsize, tick label style={font=\tiny},
      label style={font=\tiny}, axis line style={draw=clpgray!80},
      tick style={draw=clpgray!80}, ylabel shift=2pt},
    fidreal/.style={black, solid, very thick, mark=none},
    fidgan/.style={fidblue, dashed, thick, mark=none},
    fidwm/.style={fidorange, densely dotted, very thick, mark=none},
  }
  \begin{axis}[fidaxis, width={2*\hh cm}, at={(0cm,-4.62cm)},
    xlabel={true state}, ylabel={estimated state}, ytick=\empty,
    xmin=-10.9985, xmax=10.9780,
    legend pos=north west, legend cell align=left,
    legend style={font=\tiny, draw=none, fill=white, fill opacity=0.75,
                  text opacity=1, inner sep=1pt}]
    \addplot[clpgray!70, thin, dash dot, samples=2, forget plot,
             domain=-10.9985:10.9780] {x};
    \addplot[name path=rlo, draw=none, forget plot]
      table[x=x_center, y=real_q25, col sep=comma] {figures/fidelity/plots/taxinet_p.csv};
    \addplot[name path=rhi, draw=none, forget plot]
      table[x=x_center, y=real_q75, col sep=comma] {figures/fidelity/plots/taxinet_p.csv};
    \addplot[black, fill opacity=0.10, forget plot] fill between[of=rlo and rhi];
    \addplot[fidreal]
      table[x=x_center, y=real_median, col sep=comma] {figures/fidelity/plots/taxinet_p.csv};
    \addlegendentry{Held-out}
    \addplot[name path=glo, draw=none, forget plot]
      table[x=x_center, y=gan_q25, col sep=comma] {figures/fidelity/plots/taxinet_p.csv};
    \addplot[name path=ghi, draw=none, forget plot]
      table[x=x_center, y=gan_q75, col sep=comma] {figures/fidelity/plots/taxinet_p.csv};
    \addplot[fidblue, fill opacity=0.14, draw=fidblue!75, dashed,
      line width=0.3pt, forget plot] fill between[of=glo and ghi];
    \addplot[fidgan]
      table[x=x_center, y=gan_median, col sep=comma] {figures/fidelity/plots/taxinet_p.csv};
    \addlegendentry{GAN}
    \addplot[name path=wlo, draw=none, forget plot]
      table[x=x_center, y=wm_q25, col sep=comma] {figures/fidelity/plots/taxinet_p.csv};
    \addplot[name path=whi, draw=none, forget plot]
      table[x=x_center, y=wm_q75, col sep=comma] {figures/fidelity/plots/taxinet_p.csv};
    \addplot[fidorange, fill opacity=0.14, draw=fidorange!85, densely dotted,
      line width=0.4pt, forget plot] fill between[of=wlo and whi];
    \addplot[fidwm]
      table[x=x_center, y=wm_median, col sep=comma] {figures/fidelity/plots/taxinet_p.csv};
    \addlegendentry{World model}
    \draw[clpgray!90, thick] ({axis cs:-8.0478,0}|-{rel axis cs:0,1}) -- ++(0,-3pt);
    \draw[clpgray!90, thick] ({axis cs:3.5416,0}|-{rel axis cs:0,1}) -- ++(0,-3pt);
  \end{axis}
  \begin{axis}[fidaxis, width={4*\h cm}, at={(\ax cm,-4.62cm)},
    xlabel={true state}, ytick=\empty,
    xmin=0.0103, xmax=44.0]
    \addplot[clpgray!70, thin, dash dot, samples=2, forget plot,
             domain=0.0103:44.0] {x};
    \addplot[name path=alo, draw=none, forget plot]
      table[x=x_center, y=real_q25, col sep=comma] {figures/fidelity/plots/aebs_gray_d.csv};
    \addplot[name path=ahi, draw=none, forget plot]
      table[x=x_center, y=real_q75, col sep=comma] {figures/fidelity/plots/aebs_gray_d.csv};
    \addplot[black, fill opacity=0.10, forget plot] fill between[of=alo and ahi];
    \addplot[fidreal]
      table[x=x_center, y=real_median, col sep=comma] {figures/fidelity/plots/aebs_gray_d.csv};
    \addplot[name path=bglo, draw=none, forget plot]
      table[x=x_center, y=gan_q25, col sep=comma] {figures/fidelity/plots/aebs_gray_d.csv};
    \addplot[name path=bghi, draw=none, forget plot]
      table[x=x_center, y=gan_q75, col sep=comma] {figures/fidelity/plots/aebs_gray_d.csv};
    \addplot[fidblue, fill opacity=0.14, draw=fidblue!75, dashed,
      line width=0.3pt, forget plot] fill between[of=bglo and bghi];
    \addplot[fidgan]
      table[x=x_center, y=gan_median, col sep=comma] {figures/fidelity/plots/aebs_gray_d.csv};
    \addplot[name path=bwlo, draw=none, forget plot]
      table[x=x_center, y=wm_q25, col sep=comma] {figures/fidelity/plots/aebs_gray_d.csv};
    \addplot[name path=bwhi, draw=none, forget plot]
      table[x=x_center, y=wm_q75, col sep=comma] {figures/fidelity/plots/aebs_gray_d.csv};
    \addplot[fidorange, fill opacity=0.14, draw=fidorange!85, densely dotted,
      line width=0.4pt, forget plot] fill between[of=bwlo and bwhi];
    \addplot[fidwm]
      table[x=x_center, y=wm_median, col sep=comma] {figures/fidelity/plots/aebs_gray_d.csv};
    \draw[clpgray!90, thick] ({axis cs:4.1055,0}|-{rel axis cs:0,1}) -- ++(0,-3pt);
    \draw[clpgray!90, thick] ({axis cs:7.2040,0}|-{rel axis cs:0,1}) -- ++(0,-3pt);
    \draw[clpgray!90, thick] ({axis cs:14.5183,0}|-{rel axis cs:0,1}) -- ++(0,-3pt);
    \draw[clpgray!90, thick] ({axis cs:32.9606,0}|-{rel axis cs:0,1}) -- ++(0,-3pt);
  \end{axis}
\end{tikzpicture}
\vspace{0.9em}
\caption{Held-out frames and each surrogate's rendering at the same state.
The GAN row shows the released generator of each benchmark at zero latent,
and the world model row shows our model at the center of its latent box.
The plots below show the state the perception network estimates from each
source's frames (crosstrack position for taxiing, distance for AEBS)
against the true state, as a median and a 25--75\% band over the held-out
set.}
\label{fig:fidelity}
\end{figure}

%% file: figures/cellmap.tex
\definecolor{cmsafe}{RGB}{203,221,208}
\definecolor{cmfals}{HTML}{FCFF6C}
\definecolor{cmcert}{RGB}{31,86,148}
\definecolor{cmlemma}{RGB}{214,160,42}
\definecolor{cmwit}{RGB}{140,25,20}
\definecolor{cmfwd}{HTML}{B0D0D3}
\definecolor{cmalpha}{HTML}{436436}
\definecolor{cmsym}{HTML}{C44536}
\definecolor{cmback}{HTML}{D8A47F}
\definecolor{cmopen}{HTML}{8D8A86}
\definecolor{cmfalsline}{HTML}{FCFF6C}
\definecolor{cmopenline}{HTML}{8D8A86}
\begin{figure}[t]
\centering
\newcommand{\lgd}[4]{%
  \fill[#3, draw=clpgray!60, line width=0.2pt] (#1,#2) rectangle ++(0.24,0.24);
  \node[font=\tiny, text=clpgray!95, anchor=west] at (#1+0.32,#2+0.12) {#4};}
\newcommand{\exring}[2]{%
  \draw[white, line width=1.6pt] (axis cs:#1,#2) circle [radius=1.8];
  \draw[black, line width=0.7pt] (axis cs:#1,#2) circle [radius=1.8];}
\newcommand{\exshot}[4]{%
  \IfFileExists{#4}
    {\node[inner sep=0pt, draw=#3, line width=1.8pt] at (#1,#2)
       {\includegraphics[width=0.92cm, height=0.92cm]{#4}};}
    {\node[inner sep=0pt, draw=#3, line width=1.8pt, fill=clpgray!10,
           minimum width=0.92cm, minimum height=0.92cm] at (#1,#2)
       {\tiny\textcolor{clpgray}{pending}};}}
\begin{tikzpicture}
\pgfplotsset{
  cellaxis/.style={scale only axis, anchor=north west, width=5.9cm,
    height=2.95cm, xmin=0, xmax=60, ymin=0, ymax=30,
    xtick={0,10,...,60}, ytick={0,10,20,30}, axis on top,
    font=\scriptsize, tick label style={font=\tiny},
    label style={font=\tiny}, title style={font=\scriptsize, yshift=-2pt},
    axis line style={draw=clpgray!80}, tick style={draw=clpgray!80},
    xlabel={distance $d$ (m)}},
}
\begin{axis}[cellaxis, at={(0cm,0cm)},
  title={cGAN (grayscale)}, ylabel={speed $v$ (m/s)}]
  \addplot[forget plot] graphics[xmin=0, xmax=60, ymin=0, ymax=30]
    {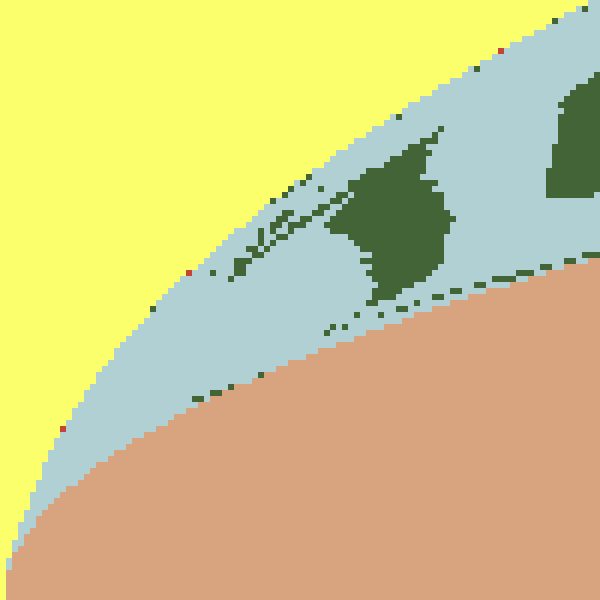};
  \exring{14.7}{9.15}
  \exring{7.5}{2.85}
  \exring{33.9}{18.75}
  \exring{6.3}{8.55}
  \exring{7.5}{18.15}
\end{axis}
\begin{axis}[cellaxis, at={(6.9cm,0cm)},
  title={World model (RGB)}, yticklabels={}]
  \addplot[forget plot] graphics[xmin=0, xmax=60, ymin=0, ymax=30]
    {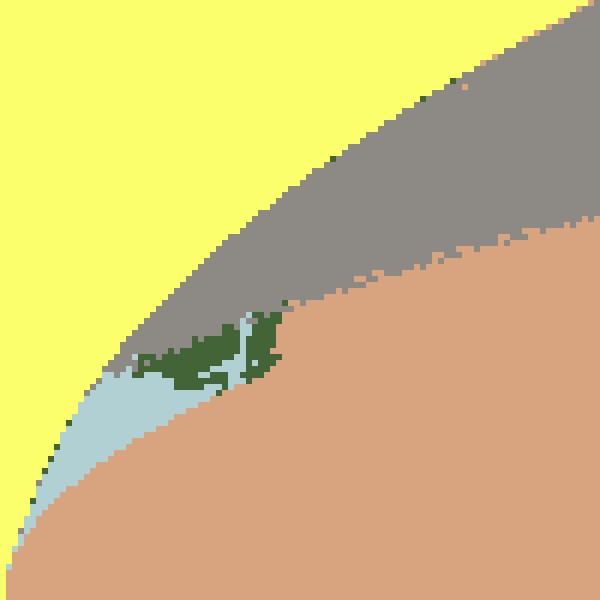};
  \exring{15.3}{9.15}
  \exring{28.5}{12.75}
  \exring{6.9}{8.85}
  \exring{7.5}{18.15}
  \exring{42.3}{18.15}
\end{axis}
\def\exdir{figures/verification/exemplars}
\exshot{0.58}{-4.20}{cmfwd}{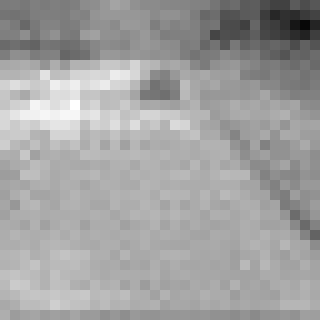}
\exshot{1.77}{-4.20}{cmback}{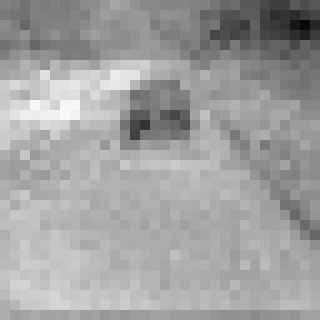}
\exshot{2.96}{-4.20}{cmalpha}{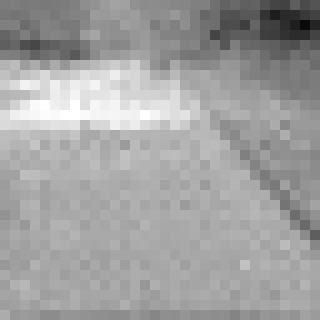}
\exshot{4.15}{-4.20}{cmsym}{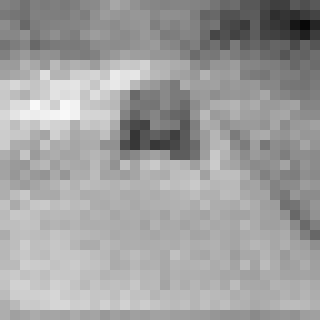}
\exshot{5.34}{-4.20}{cmfalsline}{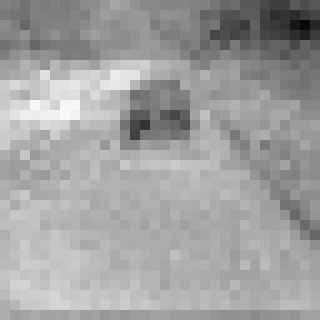}
\exshot{7.48}{-4.20}{cmfwd}{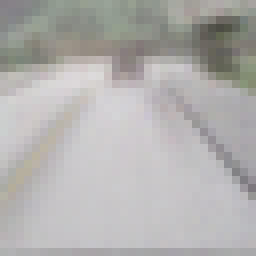}
\exshot{8.67}{-4.20}{cmback}{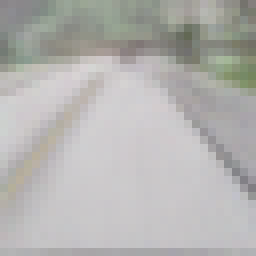}
\exshot{9.86}{-4.20}{cmalpha}{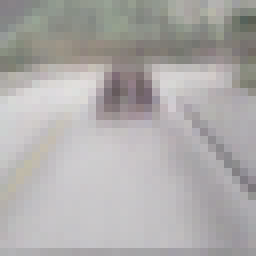}
\exshot{11.05}{-4.20}{cmfalsline}{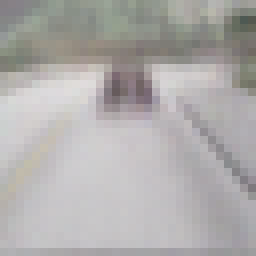}
\exshot{12.24}{-4.20}{cmopenline}{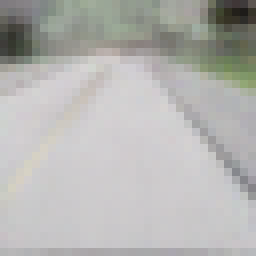}
\lgd{1.15}{-5.05}{cmfals}{falsified}
\lgd{1.15}{-5.47}{cmfwd}{forward}
\lgd{4.45}{-5.05}{cmalpha}{abstraction optimization}
\lgd{4.45}{-5.47}{cmsym}{symbolic}
\lgd{8.75}{-5.05}{cmback}{backward}
\lgd{8.75}{-5.47}{cmopen}{unresolved}
\end{tikzpicture}
\vspace{0.6em}
\caption{Each cell of the AEBS benchmark, colored by the analysis that
decided it, on the released grayscale cGAN and on our RGB world model.
Abstraction optimization includes adaptive refinement, and backward analysis
decides every cell the baseline verifies. Most unresolved cells on the world
model have not yet been analyzed. Each circle marks
one cell per outcome, rendered below by that surrogate and framed in the
outcome's color.}
\label{fig:cellmap}
\end{figure}

%% file: appendix.tex
\section{Neural Feedback System Definitions}
\label{app:nfs}
For each case study we provide the tuple
$\dynsys = \langle m, n, d, \init, \trans, \noise, \exo, \sens, \ctrl, B,
\timestep, \horizon, \reach, \avoid \rangle$ from Section~\ref{sec:bg-nfs}.
Both are discrete-time systems with step $\timestep$, and we write out
the closed-loop step of Equation~\ref{eq:next-state} for each. Where the
controller applies a fixed transform to the output of its perception
network, we fold the transform into $\ctrl$, so that $\ctrl : \R^d \to \R^m$
maps the observation to the applied control, and $B$ records only which
coordinate the control drives.

\paragraph{Aircraft Taxiing.}
The state $\vs = (p, \theta)$ is the crosstrack position (m) and
heading error (rad) of the aircraft relative to the runway
centerline~\citep{katz2022verification}. We have $n = 2$, $m = 1$, and
\[
  \trans(\vs) = (v \sin\theta,\; 0),
\]
with taxi speed $v = 5$\,m/s. The control is the heading rate
$\frac{v}{L}\tan\phi$, with wheelbase $L = 5$\,m and steering angle $\phi$,
and enters through $B = (0, 1)^\top$. The observations are from a wing-mounted camera, whose images are cropped,
converted to grayscale, and downsampled to $8 \times 16$ pixels, so the
observation dimension is $d = 128$. The set $\exo$ covers variation between
frames at a fixed state, such as downtrack position. The controller
$\ctrl(\vo) = \frac{v}{L}\tan(-0.74\,\hat p - 0.44\,\hat\theta)$ composes the
perception network (TinyTaxiNet), which estimates $(\hat p, \hat\theta)$ from
$\vo$, with a proportional steering law. The remaining components are
$\noise = \{0\}^2$, $\timestep = 0.05$, $\horizon = 320$ (a $16$\,s window),
$\init = [-10, 10] \times [-10^\circ, 10^\circ]$, $\reach = \R^2$ (so only
the avoid property constrains the system), and
$\avoid(t) = \{\vs : |p| > 10\}$, the edges of the runway. The closed-loop
step is
\[
  \nxt^{\dynsys}(\vs) = \Bigl\{ \bigl(p + v\timestep \sin\theta,\;
  \theta + \tfrac{v}{L}\timestep \tan\phi(\sens(\vs, \vxi))\bigr)
  \;\Big|\; \vxi \in \exo \Bigr\},
\]
where $\phi(\vo) = -0.74\,\hat p - 0.44\,\hat\theta$. The benchmark
updates $\phi$ at $1$\,Hz and holds it between updates. The cGAN of
\citet{katz2022verification} is conditioned on $\vs$ and carries two
latents in $[-0.8, 0.8]$.

\paragraph{Automatic Emergency Braking.}
The state $\vs = (d, v)$ is the distance to the obstacle (m) and the
speed of the vehicle (m/s)~\citep{cai2025scalable}. We have $n = 2$,
$m = 1$, and
\[
  \trans(\vs) = (-v,\; -a_0),
\]
with $a_0 = 2.53$\,m/s$^2$ the deceleration at zero braking. The braking
command $u \in [0, 1]$ enters through $B = (0, -a_1)^\top$ with
$a_1 = 5.40$\,m/s$^2$. Both constants are read from the plant layer of the
released model. The observation is a $32 \times 32$ grayscale or RGB CARLA camera frame, combined with the speed $v$, so the observation has
$1025$ or $3073$ entries. The set $\exo$ covers lighting and weather. The controller
$\ctrl(\vo) = \pi(\hat d, v)$ composes a perception head, which estimates the
distance $\hat d$ from the frame, with the released DDPG actor $\pi$, two
ReLU layers of $400$ and $300$ units with output clamped to $[0, 1]$. The
remaining components are $\noise = \{0\}^2$, $\timestep = 0.05$,
$\horizon = 238$ (an $11.9$\,s window), $\init = [0, 60] \times [0, 30]$ partitioned
into a $100 \times 100$ grid of cells verified separately,
$\reach = \{\vs : v \leq 0\}$, and $\avoid(t) = \{\vs : d \leq 0\}$. The
closed-loop step is
\[
  \nxt^{\dynsys}(\vs) = \Bigl\{ \bigl(d - v\timestep,\;
  v - (a_0 + a_1\,\ctrl(\sens(\vs, \vxi)))\,\timestep\bigr)
  \;\Big|\; \vxi \in \exo \Bigr\}.
\]
The cGAN of \citet{cai2025scalable} is conditioned on $d$ and carries four
latents in $[-0.01, 0.01]$, redrawn at every step.

\section{World Model Training}
\label{app:training}
\paragraph{Data.}
For the RGB AEBS benchmark we render frames in CARLA 0.9.16 (Town01) while
the released DDPG controller acts on the true state. We collect 240
rollouts at 20\,Hz, 13{,}056 frames in total, from initial states
stratified over $d \in [5, 60]$ and $v \in [2, 30]$. The latent $\vz$ is
the sun altitude, drawn from $[0^\circ, 90^\circ]$ once per rollout. Frames
are cropped, resized to $32 \times 32$, and scaled to $[-1, 1]$. The state
and latent are normalized by their ranges. We split the data by rollout
into training, validation, and held-out sets of 150, 40, and 50 rollouts,
stratified by $\vz$, so that no held-out image comes from a trajectory seen
in training.

\paragraph{Architecture and optimization.}
The linear layer of Equation~\ref{eq:decoder} lifts the state to a
$32 \times 4 \times 4$ map, followed by $L = 3$ stages with 32, 16, and 8
channels and a $3 \times 3$ output convolution. The model has 28{,}591
parameters. We minimize the loss of Section~\ref{sec:swm} with Adam
(learning rate $2 \times 10^{-4}$, weight decay $10^{-4}$, gradient-norm
clipping at $1.0$, batch size 256) for 120 epochs without early stopping,
and keep the final weights. Training takes about 30\,s on one H100.

\paragraph{Perception heads.}
The perception heads for the RGB benchmark are trained on the CARLA frames,
with the same split, an $\ell_1$ loss on $d / 60$, and the same optimizer
settings for 60 epochs.

\section{Fidelity Evaluation}
\label{app:fidelity}
Each surrogate renders an image at the recorded state of every held-out
frame. The RGB world model also receives the frame's sun altitude. GAN
latents are set to the center of their range, and world models without a
physical latent are rendered at the center of their latent box. All images
are mapped to $[0, 1]$ before scoring. RMSE is taken over all pixels and
channels. SSIM uses a $7 \times 7$ uniform window over full windows, with
$K_1 = 0.01$ and $K_2 = 0.03$, and is averaged over channels for RGB.
Generator outputs outside $[0, 1]$ are clipped for SSIM only.

\paragraph{Emergency braking.}
The 2{,}487 held-out frames are the frames of the held-out rollouts of the
RGB world model's split. The cGAN and SAGAN are the generators released by
\citet{cai2025scalable}. Their training data is not published, so we cannot
confirm that these frames are held out from them.

\paragraph{Aircraft taxiing.}
We use the 10{,}000 downsampled frames released by
\citet{katz2022verification}, split 80/10/10 and stratified on position and
heading deciles, which leaves 1{,}002 held-out frames. Our world models
train on the 7{,}996 training frames, with hyperparameters selected on the
validation split. The DCGAN weights of \citet{katz2022verification} were
not released, so we retrain the DCGAN from their code on the same training
frames (binary cross-entropy loss, batch size 256, Adam with learning rate
$7 \times 10^{-4}$ and $(\beta_1, \beta_2) = (0.5, 0.99)$, 750 epochs, two
latents). The MLP GAN is their released generator. It was distilled from a
DCGAN trained on all 10{,}000 frames, so it has seen our held-out frames
in training.

\section{Verification Procedure Settings}
\label{app:settings}
Table~\ref{tab:procedure-settings} lists the settings of
Algorithm~\ref{alg:procedure} for both AEBS experiments. Each cell of the
$100 \times 100$ grid is $0.6$\,m by $0.3$\,m/s. Refinement splits a cell
into $K \times K$ sub-cells, and the cell is verified only when all $K^2$
sub-cells are. An analysis fails on a sub-cell at the first step whose
distance lower bound is non-positive, and verifies it at the first step
whose speed upper bound is non-positive. A sub-cell that reaches the
horizon without either remains unresolved. At $K \in \{4, 6\}$ a pass stops at the first
failing sub-cell. At $K \in \{1, 2\}$ every sub-cell is analyzed so that
verified pieces can be reused.

\begin{table}[h]
\centering
\small
\caption{Procedure settings for the grayscale (cGAN) and RGB (world model)
experiments. ``Same'' means the cGAN setting.}
\label{tab:procedure-settings}
\begin{tabular}{lll}
\toprule
Setting & cGAN & World model \\
\midrule
Falsification & 5{,}000 samples per cell & same \\
Targeted falsification & 4{,}000 samples, 5 rounds & same \\
Refinement $K$ & $\{1, 2, 4, 6\}$; 12 for three cells & $\{1, 2, 4, 6\}$ \\
Abstraction optimization & 20 iterations, then 5 per step; rate 0.5 & same \\
Neuron splitting & off & 16--64 domains; 300\,s per cell \\
Symbolic analysis & 3{,}600\,s budget & 120\,s budget \\
Backward analysis & from the last forward step & same; 300\,s budget \\
\bottomrule
\end{tabular}
\end{table}

\paragraph{Numerical soundness.}
Bounds are computed in float32 on the GPU with TF32 disabled, so matrix
products keep full float32 precision. The sign tests that decide each
sub-cell are done in float64. We do not claim soundness against
floating-point error inside the bounding library. We record discrepancies between float32 and float64 to detect floating point errors in the bounding library.

\paragraph{Perception-head budget.}
The three heads of Table~\ref{tab:budget} read the same RGB world model and
verify the same $10 \times 10$ block of cells, $d \in [10.8, 16.8]$ and
$v \in [8.7, 11.7]$. No cell in the block is falsified or verified before
the experiment. Cells are analyzed in order of
decreasing clearance, and a head that finishes the block continues on the
remaining unresolved cells in the same order. GPU-hours are the sum of
per-cell wall-clock times. Up to three processes shared one H100, so these
times include contention between them. A head stops once
its in-flight time reaches 10 hours, and a cell cut off mid-flight
contributes neither a verdict nor time. 

\section{Verification Breakdown by Analysis}
\label{app:methods}
Tables~\ref{tab:cgan-methods} and~\ref{tab:wm-methods} attribute each
of the 10{,}000 cells to the analysis of Section~\ref{sec:procedure} that
decided it.

\begin{table}[h]
\centering
\small
\caption{Cells of the grayscale AEBS benchmark decided by each analysis,
on the released cGAN surrogate. Backward analysis decides all 2{,}669
cells that \citet{cai2025scalable} verify.}
\label{tab:cgan-methods}
\begin{tabular}{lrr}
\toprule
Analysis & Verified & Falsified \\
\midrule
Falsification & & 3{,}464 \\
Forward analysis & 2{,}208 & \\
Abstraction optimization and refinement & 551 & \\
Symbolic analysis & 3 & \\
Backward analysis & 3{,}774 & \\
\midrule
Total & 6{,}536 & 3{,}464 \\
\bottomrule
\end{tabular}
\end{table}

\begin{table}[h]
\centering
\small
\caption{Cells of the RGB AEBS benchmark decided by each analysis, on the
world-model surrogate. Of the 1{,}825 unresolved cells, 1{,}822 have not
yet been analyzed.}
\label{tab:wm-methods}
\begin{tabular}{lrr}
\toprule
Analysis & Verified & Falsified \\
\midrule
Falsification & & 3{,}498 \\
Forward analysis & 305 & \\
Abstraction optimization and refinement & 156 & \\
Symbolic analysis & 0 & \\
Backward analysis & 4{,}216 & \\
\midrule
Total & 4{,}677 & 3{,}498 \\
\bottomrule
\end{tabular}
\end{table}